\documentclass[conference]{IEEEtran}
\IEEEoverridecommandlockouts
\usepackage{cite}
\usepackage{amsmath,amssymb,amsfonts}
\usepackage{algorithmic}
\usepackage{graphicx}
\usepackage{textcomp}
\usepackage{xcolor}
\input{mysymbol.sty}

\renewcommand\baselinestretch{.97}
    
\begin{document}

\title{Generalizing Graph Convolutional Neural Networks with Edge-Variant Recursions on Graphs\\
\thanks{The authors are with the Department of Electrical \& Systems Engineering, University of Pennsylvania, 19104 Philadelphia, United States. This work is supported by NSF CCF 1717120, ARO W911NF1710438, ARL DCIST CRA W911NF-17-2-0181, ISTC-WAS and Intel DevCloud. E-mails: {$\{$eisufi, fgama, aribeiro$\}$@seas.upenn.edu}.}
}

\author{\IEEEauthorblockN{Elvin Isufi,\, Fernando Gama,\, and Alejandro Ribeiro}
\vspace{-2cm}}

\maketitle

\begin{abstract}
This paper reviews graph convolutional neural networks (GCNNs) through the lens of edge-variant graph filters. The edge-variant graph filter is a finite order, linear, and local recursion that allows each node, in each iteration, to weigh differently the information of its neighbors. By exploiting this recursion, we formulate a general framework for GCNNs which considers state-of-the-art solutions as particular cases. This framework results useful to \emph{i)} understand the tradeoff between local detail and the number of parameters of each solution and \emph{ii)} provide guidelines for developing a myriad of novel approaches that can be implemented locally in the vertex domain. One of such approaches is presented here showing superior performance w.r.t. current alternatives in graph signal classification problems.
\end{abstract}\vskip-.2cm

\begin{IEEEkeywords}
Graph convolutional neural networks; graph signal processing; graph filters; edge-variant.
\end{IEEEkeywords}\vspace{-.2cm}

\section{Introduction}\vskip-2mm

Graph convolutional neural networks (GCNNs) are gaining momentum as a promising tool for addressing a variety of classification and regression tasks for data that live in irregular spaces, such as network data \cite{scarselli2009graph}. With the aim to replicate the success of traditional CNNs, GCNNs play a fundamental role in semi-supervised learning on graphs and classifying different graph signals (i.e., values indexed by the nodes of the graph).

A central aspect of GCNNs is the extension of the convolution operation to graph signals. The seminal work \cite{bruna2013spectral} defined convolution as the point-wise multiplication in the graph spectral domain between the projected graph signal in this domain and the transfer function of a learnable filter. Such an approach finds solid grounds on graph signal processing (GSP) theory \cite{ortega2018graph}, where the graph spectrum plays the role of the Fourier basis for graph signals. Subsequently, several works exploited this connection and proposed computationally lighter GCNN models. In particular, \cite{kipf2016semi,defferrard2016convolutional, du2017topology, gama2018convolutional} used the so-called polynomial graph filters \cite{ortega2018graph}, while \cite{levie2017cayleynets,bianchi2019graph} relied on graph filters having a rational transfer function \cite{isufi2017autoregressive}. Differently, \cite{gama2018convolutionalNV} designed architectures by using the node-variant (NV) graph filters \cite{segarra2017optimal}, while \cite{gama2018mimo} introduced MIMO approaches to learn from multiple features.

While the above works introduce techniques that extend CNNs to graphs, they are mostly based on (well-motivated) analogies with classical neural networks. Such a strategy presents, however, its own limitations towards extending these methods to more involved ones that better exploit the graph.

The main aim of this paper is to formulate a general framework that unifies state-of-the-art GCNN architectures facilitating comparison, showing their limitations, and highlighting their potential. For such a goal, it explores the so-called edge-variant (EV) graph filters \cite{Coutino2017,coutino2018advances}. The EV graph filter is a local, linear, and finite-order recursion in the node domain where each node weighs differently, in each iteration, the information in its neighborhood. Therefore, it presents the most general linear and local operation that a node can do --gather information from all neighbors and weight each of them differently. This local property puts the EV as a computationally efficient candidate (only local information is exchanged) for capturing detail at the node connection level.

Nevertheless, in the general form, the learnable parameters of the EV graph filter depend on the number of graph edges. To tackle the latter, we first show how state-of-the-art solutions fall under the EV recursion and how they impose parsimonious models on the learnable parameters, rendering their number independent of the graph dimensions. Then, we explore such insights to provide guidelines for designing a variety of novel architectures in the spirit of the EV recursion whose number of parameters is independent of the graph dimensions.

In a nutshell, the contributions of this paper are: $i)$ To formulate a general framework for GCNNs through EV graph filters. $ii)$ To show how the state-of-the-art approaches are specific parameterizations of this EV recursion. $iii)$ To present rigorous design guidelines for GCNNs based on the EV recursion that preserve locality and whose number of parameters is independent of the graph dimension. $iv)$ To introduce one new such an architecture and show its superior performance for graph signal classification tasks.


\section{Background}\vskip-1mm

\subsection{Graphs and graph filters}

Let $\ccalG \!=\! \{\ccalV, \ccalE\}$ be a weighted graph with vertex set $\ccalV \!=\! \{1, \ldots, N\}$ of cardinality $|\ccalV| \!=\! N$ and edge set $\ccalE \subseteq \ccalV\times \ccalV$ composed of $|\ccalE| \!\!=\!\! M$ ordered pairs $(i,j) \!\!\in\! \ccalE$ iff there is an edge between nodes $i$ and $j$. For each node $i$, we define the neighborhood set $\ccalN_i \!=\! \{j\!:\!(j,i) \!\in\! \ccalE\}$ as the set of nodes connected to $i$. The sparsity of the edge set of $\ccalG$ is represented by an $N\!\times\! N$ matrix $\bbS$, named the graph shift operator matrix, where $[\bbS]_{i,j} \!\neq\! 0$ if $(j,i) \!\in\! \ccalE$ or $i \!=\! j$. Candidates for $\bbS$ are the graph adjacency $\bbW$, the graph Laplacian matrix (undirected graphs) or any of their normalizations. For generality, in the sequel, we will focus on directed graphs.

Along with $\ccalG$, consider a set of signal values (features) $\bbx = [x_1, \ldots, x_N]^T \in \reals^N$ in which component $x_i$ resides on node $i$. 
By exploiting the coupling between the graph $\ccalG$ and the graph signal $\bbx$, it is possible to compute a graph harmonic analysis for $\bbx$ similarly to the one performed for temporal and image signals. Specifically, given the eigendecomposition $\bbS = \bbU\bbLambda\bbU^{-1}$, the graph Fourier transform (GFT) of $\bbx$ is $\hbx = \bbU^{-1}\bbx$.
Likewise, the inverse transform is $\bbx = \bbU\hbx$. Here, $\bbU$ contains along the columns the oscillating modes of the graph and $\hbx$ are the respective Fourier coefficients. The eigenvalues $\bbLambda$ represent the spectral support for $\hbx$ and are commonly referred to as the graph frequencies \cite{ortega2018graph}.

Given the Fourier expansion, we can now filter $\bbx$ directly in the spectral domain. That is, for $h: \mbC \to \reals$ being the filter spectral response (transfer function), the filter output $\hby = h(\bbLambda)\hbx$
%
%
is computed as the convolution (pointwise multiplication in the spectral domain) between the filter transfer function $h(\bbLambda)$ and the GFT of $\bbx$. By means of the inverse GFT, the vertex domain output becomes
\begin{equation}\label{eq.filtVx}
\bby = \bbH(\bbS)\bbx \quad \text{and}\quad\bbH(\bbS) = \bbU h(\bbLambda)\bbU^{-1}.
\end{equation}

However, \eqref{eq.filtVx} is not local, since in computing the output $y_i$, node $i$ needs access to the graph signal of non-neighboring nodes. To account for the locality, we can define the filtering operation directly in the vertex domain as the aggregation of neighboring information. The node $i$ output for an order one local filter is\vskip-3mm
\begin{equation}\label{eq:vxFilt}
y_i = \sum_{j \in \ccalN_i \cup i}\phi_{i,j}x_j
\end{equation}
where the scalar parameter $\phi_{i,j}$ weighs the information of the neighboring node $j$. Nevertheless, this direct vertex domain definition does not enjoy a spectral behavior analysis limiting the connection with the convolution operation.

One way to link the spectral and the vertex domain filtering is to consider the polynomial graph filters~\cite{ortega2018graph} with output
\begin{align}\label{eq:pol}
\begin{split}
\bby &= \phi_0\bbI_N\bbx + \phi_1\bbS\bbx + \ldots + \phi_K\bbS^K\bbx\\
&\triangleq \phi_0\bbw^{(0)} + \phi_1\bbw^{(1)} + \ldots + \phi_K\bbw^{(K)}.
\end{split}
\end{align}
Due to the locality of $\bbS$, $\bby$ can be obtained in the vertex domain through local information exchange. The state $\bbw^{(0)}$ is simply the graph signal, while $\bbw^{(1)} = \bbS\bbx$ consists of one-hop information exchange between adjacent nodes. The higher order states $\bbw^{(k)} = \bbS^k\bbx$ are computed recursively as $\bbw^{(k)} = \bbS\bbw^{(k-1)}$, i.e., by exchanging with the neighbors the previous intermediate state $\bbw^{(k-1)}$. Such an implementation amounts for a complexity of order $\ccalO(MK)$. By means of the GFT, the filtering operation in \eqref{eq:pol} has the transfer function\vskip-2mm
\begin{equation}\label{eq:polyResp}
h(\bbLambda) = \sum_{k = 0}^K\phi_k\bbLambda^k.
\end{equation}
Therefore, we conclude that the output $\bby$ in \eqref{eq:pol} consists of the convolution between a graph filter with a polynomial transfer function and $\bbx$. This filter enjoys a local implementation and captures detail in a neighborhood of radius $K$ from the node.\vspace{-1mm}

\subsection{Edge-variant graph filters}\vskip-1mm

The edge-variant graph filter is a finite order recursion implemented in the vertex domain in the form similar to \eqref{eq:vxFilt} \cite{coutino2018advances}. Let $\{\bbPhi^{(k)}\}_{k=1}^K \in \reals^{N\times N}$ be a collection of $K$ matrices that share the sparsity pattern of $\bbI_N + \bbS$. The intermediate states of the EV filter are computed recursively as $\bbw^{(1)} = \bbPhi^{(1)}\bbx$; $\bbw^{(2)} = \bbPhi^{(2)}\bbw^{(1)} = \bbPhi^{(2)}\bbPhi^{(1)}\bbx$; and
\begin{equation}\label{eq:stateEV}
\bbw^{(k)} = \bbPhi^{(k)}\bbw^{(k-1)} =  \bbPhi^{(k)} \bbPhi^{(k-1)}\ldots \bbPhi^{(2)} \bbPhi^{(1)}\bbx.
\end{equation}
%
Since $\bbPhi^{(1)}$ shares the support with $\bbS + \bbI_N$, the state $\bbw^{(1)}$ accounts also for the scaling of $\bbx$ through the diagonal elements (i.e., each node $i$ scales its own signal with a different parameter $\phi_{i,i}^{(1)}$). The higher order states $\bbw^{(k)}$ for $k > 1$ are again obtained recursively since the parameter matrices $\bbPhi^{(k)}$ respect the graph connectivity. Put differently, each $\bbPhi^{(k)}$ considers a different parameter $\phi_{i,j}^{(k)}$ for each edge $(j,i)\in \ccalE$ and adds potential self-loops through $\phi_{i,i}^{(k)}$. Node $i$ computes then the $k$th order state as
\begin{equation}
w^{(k)}_i = \sum_{j \in \ccalN_i \cup i}\phi_{i,j}^{(k-1)}w^{(k-1)}_j.
\end{equation}
%

By defining $\bbPhi^{(k:1)} = \prod_{\kappa = 1}^k\bbPhi^{(\kappa)}$ and  putting back together all terms in \eqref{eq:stateEV}, the output of an order $K$ EV graph filter is
\begin{equation}\label{eq:EVout}
\bby = \sum_{k = 0}^K\bbPhi^{(k:1)}\bbx = \overbrace{\sum_{k = 1}^K\left(\prod_{\kappa = 1}^k\bbPhi^{(\kappa)}	\right)}^{\bbH(\bbS)} \bbx.
\end{equation}
%
%
The total number of parameters of the edge-variant graph filter is $K(M+N)$, which is in general smaller than the $N^2$ parameters of an arbitrary linear transform $\bby = \bbH\bbx$. In computing the output $\bby$ in \eqref{eq:EVout} the EV graph filter incurs in an overall computational complexity of order $\ccalO(K(M+N))$ which is similar to that of \eqref{eq:pol} since in general $M \approx N$ or we can consider an EV recursion without self-loops.

The ability to capture local detail at the edge level and the reduced implementation complexity is leveraged next to define graph neural networks (GNN) with a controlled number of parameters and computational complexity matched to the sparsity pattern of the graph.

\section{Edge-variant graph neural networks}
\label{sec:main}

Consider a training set $\ccalT = \{\bbx_i, \bbz_i\}_{i = 1}^{|\ccalT|}$ composed of $|\ccalT|$ examples of inputs $\bbx_i \in \ccalX$ and output representations $\bbz_i \in \ccalZ$. A GNN leverages the underlying graph representation of the data and learns a model $f(\cdot; \bbS): \ccalX \to \ccalZ$ such that $\tbz_i = f(\bbx_i;\bbS)$ minimizes some loss function $\ccalL(\tbz_i,\bbz_i)$ for $\bbz_i \in \ccalT$ and generalizes well for $\bbz_i \notin \ccalT$. 

To capture several levels of detail, model $f(\cdot, \bbS)$ is layered into the cascade of $L$ functions $f(\cdot, \bbS) = f^L(\cdot, \bbS)\circ f^{L-1}(\cdot, \bbS) \circ\ldots \circ f^1(\cdot, \bbS)$ each consisting of a succession of linear transforms and nonlinearities. Layer $l$ produces as output a collection of $F_l$ higher level signal features obtained through processing the $F_{l-1}$ features $\bbz_1^{l-1}, \ldots, \bbz_{F_{l-1}}^{l-1}$ computed at the previous layer. The $f$th higher level feature is computed as 
\begin{equation}\label{eq:gnnGen}
\bbz_f^l = \sigma^l\left(\sum_{g = 1}^{F_{l-1}}\bbH_{f,g}^l(\bbS)\bbz_g^{l-1}		\right)
\end{equation}
where $\sigma^l(\cdot)$ represents the nonlinearity that might be point-wise (e.g., ReLU) or graph dependent \cite{ruiz2018median} and $\bbH_{f,g}^l(\bbS)$ leverages the graph structure to relate the $g$th input feature $\bbz_g^{l-1}$ to the $f$th output feature $\bbz_f^l$. 

The graph basically serves as a parameterization to reduce both the computational complexity and the number of parameters. GCNNs, in particular, consider $\bbH_{f,g}^l(\bbS)$ to be a graph filter that has a spectral interpretation such as \eqref{eq.filtVx} or \eqref{eq:polyResp}. In the sequel, we consider $\bbH_{f,g}^l(\bbS)$ to be an EV graph filter and show that current approaches represent different parameterizations to induce the spectral convolution into the EV GNN. 

\subsection{Properties of the edge-variant neural network layer}

From \eqref{eq:EVout} and \eqref{eq:gnnGen}, we build the EV GNN layer as\vskip-2mm
\begin{equation}\label{eq:evConv}
\bbH_{f,g}^l(\bbS) = \sum_{k = 1}^K\bbPhi^{l,(k:1)}_{f,g}
\end{equation}
characterized by the following properties. 

First, it does not require the knowledge of $\bbS$, but only of its support. This is because, differently from current solutions, it will learn from the training data $\ccalT$ a collection of $K$ parameter matrices $\bbPhi^{(k)}$, where each of them acts as a different graph shift operator. Therefore, it represents a robust learning strategy for data residing on graphs whose edge weights are known up to some uncertainty, known only partially, or not known at all, such as biological networks \cite{wang2018network}.

Second, the computational complexity of each layer is linear in the graph parameters. By setting, $F = \max_l F_l$, the overall complexity of the EV layer is of order $\ccalO(KF^2(M+N))$ matching that of current state-of-the-art GCNN approaches. 

Third, the number of parameters per layer is, at most $KF^2(M+N)$. The latter, although allowing the EV to have the maximum degrees of freedom given a topology, may often be a limitation for large graphs or when $|\ccalT|$ is small. Our goal in the next section is, therefore, to show that GCNN layers proposed in the literature are particular cases of \eqref{eq:evConv}. Establishing these relationships allows the proposal of novel solutions that increase the descriptive power while preserving an efficient implementation complexity. 

\subsection{Parametrizations}

\textbf{Polynomial GCNN.} Several variants of GCNNs introduced in the literature use at each layer graph filters of the form
\begin{equation}\label{eq:polFilt}
\bbH(\bbS) = \sum_{k = 0}^K\phi_k\bbS^k
\end{equation}
where in the filter definition $\bbH(\bbS) = \bbH^l_{f,g}(\bbS)$ we omitted the layer and feature indices to simplify notation. This is the case of \cite{du2017topology, gama2018convolutional} which use general polynomials and \cite{kipf2016semi,defferrard2016convolutional} that consider Chebyshev polynomials.

These filters can be expressed in the form \eqref{eq:evConv} by restricting the parameter matrices to $\bbPhi^{(k:1)} \!=\! \phi_k\bbS^k$ for $k\!>\! 1$ and $\bbPhi^{(1)} \!=\! \phi_0\bbI_N \!+\! \phi_1\bbS$. In other words, the EV and the polynomial recursions represent two extremes to implement graph filters locally. The EV recursion allows each node $i$ to learn for each iteration $k$ a different parameter $\phi_{i,j}^{(k)}$ that reflects the importance of $j$'s node features to node $i$. The polynomial implementation instead forces all nodes to weigh the information of all neighbors with the same parameter $\phi_k$ within the $k$th iteration. However, this restriction makes the number of parameters $(K+1)F^2$ independent from $M$ and $N$. 

This way of parameterizing the EV recursion creates now opportunities for proposing a myriad of intermediate solutions that extend \eqref{eq:polFilt} towards the edge-variant implementation \eqref{eq:evConv}. One such an approach may be a recursion that in addition to \eqref{eq:polFilt} considers parameters also for the most critical edges (e.g., edges without which the graph becomes disconnected).

\emph{Remark 1:} Along with the above works, also \cite{simonovsky2017dynamic,monti2017geometric,atwood2016diffusion} and \cite{levie2017cayleynets,bianchi2019graph} fall under the lines of the polynomial filtering \eqref{eq:polFilt}. In specific, \cite{simonovsky2017dynamic} considers single shifts on graphs using as graph shift operator a learnable weight matrix, \cite{monti2017geometric} considers a Gaussian kernel to mix the neighboring node information, while \cite{atwood2016diffusion} uses random walks. The works in \cite{levie2017cayleynets,bianchi2019graph}, although aiming to build a GCNN layer by using graph filters with a rational transfer function, approximate the inherited matrix inverse in the vertex domain by a finite iterative algorithm. This finite approximation implicitly transforms these techniques into polynomial recursions whose order depend on the number of iterations (see also \cite{isufi2017autoregressive} for more detail). \hfill$\square$

\textbf{Spectral GCNN.} We here establish a link between the edge-variant recursion \eqref{eq:evConv} and the spectral GCNN \cite{bruna2013spectral} to provide more insights on its convolutional behavior. The spectral GCNN exploits \eqref{eq.filtVx} and learns directly the filter a transfer function $h(\bbLambda) = \diag(h(\lambda_0), \ldots, h(\lambda_{N-1}))$. To keep the number of parameters independent from $N$, $h(\bbLambda)$ is parameterized as\vskip-3mm
\begin{equation}\label{eq:specGCNN}
h(\bbLambda) = \diag(\bbB\bbb)
\end{equation}\vskip-1mm
\noindent where $\bbB \in \reals^{N\times b}$ is a prefixed kernel matrix and $\bbb \in \reals^b$ are the $b \ll N$ learnable parameters. Therefore, the number of parameters for each layer is at most $bF^2$, while the computational complexity is of order $\ccalO(N^2)$ required to compute the GFT of the features. Additionally, such an approach requires the eigendecomposition of $\bbS$ (order $\ccalO(N^3)$ to be computed once) and the learned $\bbH(\bbS) = \bbU\diag(\bbB\bbb)\bbU^{-1}$ does not capture the local detail around each vertex.

Nevertheless, this spectral interpretation is useful to understand the EV behavior. We can force the EV recursion \eqref{eq:evConv} to have a spectral response by restricting all coefficient matrices to share the eigenvectors with $\bbS$, i.e., $\bbPhi^{(k)} = \bbU\bbLambda^{(k)}\bbU^{-1}$ \cite{coutino2018advances}. Then, EV transfer function becomes\vskip-2mm
\begin{equation}\label{eq:evResp}
h(\bbLambda) = \sum_{k = 1}^K\left(\prod_{\kappa = 1}^k\bbLambda^{(k)}	\right).
\end{equation}
Subsequently, let $\ccalI$ be the index set defining the zero entries of $\bbS + \bbI_N$. The fixed support condition for each $\bbPhi^{(k)}$ is\vskip-2mm
\begin{equation}\label{eq:suppVec}
\bbC_{\ccalI}\text{vec}\left(\bbPhi^{(k)}	\right) = \bbzero_{|\ccalI|}
\end{equation}\vskip-1mm
\noindent where $\bbC_{\ccalI} \in \{0,1\}^{|\ccalI|\times N^2}$ is a selection matrix whose rows are those of $\bbI_{N^2}$ indexed by $\ccalI$, $\text{vec}(\cdot)$ denotes the vectorization operation, and $\bbzero_{|\ccalI|}$ is the ${|\ccalI|} \times 1$ vector of all zeros. From the properties of the $\text{vec}(\cdot)$ operator, \eqref{eq:suppVec} becomes\vskip-2mm
\begin{align}\label{eq.lamLin}
\begin{split}
\bbC_{\ccalI}\text{vec}\left(\bbU\bbLambda^{(k)}\bbU^H\right) = \bbC_{\ccalI}\text{vec}\left(\bbU*\bbU^H\right)\bblambda^{(k)} = \bbzero_{|\ccalI|}
\end{split}
\end{align}
where ``*" denotes the Khatri-Rao product and $\bblambda^{(k)} = \diag(\bbLambda^{(k)})$ is the $N\times 1$ vector composed by the diagonal elements of $\bbLambda^{(k)}$. Put differently, \eqref{eq.lamLin} implies\vskip-2mm
\begin{equation}
\bblambda^{(k)} \in \text{null}\left(\bbC_{\ccalI}\text{vec}\left(\bbU*\bbU^H\right)	\right).
\end{equation}
Finally, by considering $\bbB_{\bbU,\ccalI}$ as a basis that spans the nullspace of $\bbC_{\ccalI}\text{vec}\left(\bbU*\bbU^H\right)$, we can expand $\bblambda^{(k)}$ as $\bblambda^{(k)} = \bbB_{\bbU,\ccalI}\bbmu^{(k)}$ and write \eqref{eq:evResp} as\vskip-2mm
\begin{equation}
h(\bbLambda) = \sum_{k = 1}^K\left(\prod_{\kappa = 1}^k\diag\left(\bbB_{\bbU,\ccalI}\bbmu^{(k)}\right)	\right)
\end{equation}
for some basis expansion coefficients $\bbmu^{(k)}$.

The eigendecomposition of $\bbPhi^{(k)}$ implicitly reduces the total number of layer parameters from $KF^2(M+N)$ to $KF^2$rank$(\bbB_{\bbU,\ccalI})$ with rank$(\bbB_{\bbU,\ccalI}) \ll N$. That is, there is a subclass of the EV recursion that respects the operation of convolution, but, differently from \eqref{eq:specGCNN}, it captures local detail in the vertex domain and enjoys a linear implementation complexity. This subclass has also analogies with \eqref{eq:specGCNN}, which is obtained by setting $K = 1$, $\bbB_{\bbU,\ccalI} = \bbB$, and $\bbmu^{(1)} = \bbb$.

In general, we may conclude that the EV recursion implements a GNN layer that goes beyond convolution. Drawing analogies with linear system theory, the GCNN approaches behave as a linear time-invariant (now shift-invariant \cite{ortega2018graph}) filter, while the EV graph filter behaves as linear time-varying (now shift-varying; a different shift per $k$) filter that trades the convolutional interpretation with the ability to capture time-varying (now shift-varying) detail.

\textbf{Node-variant GCNN.} The idea to propose GNNs that extend convolution is also considered in \cite{gama2018convolutionalNV}, which proposed an architecture having as graph filter the recursion
\begin{equation}\label{eq:nodeV}
\bbH(\bbS) = \sum_{k = 0}^K\diag\left(\bbC_{\ccalB}\bbphi^{(k)}_\ccalB	\right)\bbS^k
\end{equation}
where $\ccalB \subset \ccalV$ is a set of \emph{privileged} nodes (e.g., the $|\ccalB|$ nodes with the highest degree), $\bbC_\ccalB \in \{0,1\}^{N \times |\ccalB|}$ is a tall binary matrix, and $\bbphi^{(k)}_\ccalB \in \reals^{|\ccalB|}$ is a vector of parameters for the nodes in $\ccalB$. In short, \eqref{eq:nodeV} learns for each shift, $|\ccalB|$ different coefficients for the nodes in $\ccalB$ and then maps them through $\bbC_\ccalB$ to the remaining nodes $\ccalV\backslash\ccalB$.

This filter is another way to restrict the EV degrees of freedom, which parameterizes the coefficient matrices to $\bbPhi^{(1)} \!=\! \diag\left(\bbC_{\ccalB}\bbphi^{(0)}_\ccalB	\right) \!+\! \diag\left(\bbC_{\ccalB}\bbphi^{(1)}_\ccalB	\right)\bbS$ and $\bbPhi^{(k:1)} \!=\! \diag\left(\bbC_{\ccalB}\bbphi^{(k)}_\ccalB	\right)\bbS^k$ for $k \!>\! 1$. That is, \eqref{eq:nodeV} is an intermediate approach between the polynomial \eqref{eq:polFilt} and the EV \eqref{eq:evConv} recursions and allows each node $i \in \ccalB$ to learn, for each $k$, a different parameter $\phi_i^{(k)}$ that reflects the importance of all its neighborhood to node $i$. The total number of parameters per layer is at most $(K+1)F^2|\ccalB|$ while the computational complexity is similar to that of \eqref{eq:polFilt}.

This different way of parameterizing the EV recursion provides alternative choices to build new intermediate architectures that lever the idea of privileged nodes while giving importance to the edge-based detail. In the sequel, we propose one such an extension that merges insights from the EV, the polynomial, and the NV architecture.

\subsection{Hybrid edge-variant neural network layer}
The hybrid edge-variant (HEV) layer considers the linear operation in \eqref{eq:gnnGen} to be a graph filter of the form
\begin{equation}\label{eq:HEV}
\bbH(\bbS) = \sum_{k = 0}^{K}\left(\prod_{\kappa = 0}^k\bbPhi_\ccalB^{(\kappa)} + \phi_k\bbS^k	\right)
\end{equation}
where $\bbPhi^{(0)}_{\ccalB} \in \reals^{N\times N}$ is a diagonal matrix whose $i$th diagonal element $\phi^{(0)}_{i,i} \neq 0$ iff node $i$ belongs to the privileged set $\ccalB \subset \ccalV$; $\{\bbPhi_\ccalB^{(k)}\}_{k = 1}^K \in \reals^{N \times N}$ are a collection of $K$ matrices whose $(i,j)$th element $\phi^{(k)}_{i,j} \neq 0$ iff $i \in \ccalB$ and $j \in \ccalN_i \cup i$; and $\{\phi_k\}_{k = 0}^K$ are a collection of $K+1$ scalars. Put simply, recursion \eqref{eq:HEV} allows nodes in $\ccalB$ to learn node-varying parameters for $k = 0$ and edge-varying parameters for $k\ge1$, while the nodes in $\ccalV\backslash\ccalB$ learn global parameters similar to \eqref{eq:polFilt}.

This approach represents yet another intermediate architecture between the full convolutional ones and the full EV. By setting $N_{\ccalB,\text{max}} = \max_{i \in \ccalB}|\ccalN_i|$ as the maximum number of neighbors for the nodes in $\ccalB$, the overall number of parameters per layer is at most $F^2((K+1) + |\ccalB| + K|\ccalB|N_{\ccalB,\text{max}})$. Finally, the HEV implementation cost is of order $\ccalO(KF^2(2M+N))$.

\section{Numerical results}
\label{sec:numRes}


We compare the proposed edge-variant and hybrid edge-variant architectures with the spectral, polynomial, and the node-variant alternatives on a source localization and an author attribution problem. For both experiments, we designed all architectures (except for the spectral GCNN) to have the same computational cost.


\subsection{Source Localization} \label{subsec:sourceLoc}

\textbf{Setup.} The goal of this experiment is to find out which community in a stochastic block model (SBM) graph is the source of a diffusion process by observing different diffused signals originated at different (unknown) communities at different (unknown) time instants. $\ccalG$ is an undirected SBM graph of $N = 50$ nodes divided equally into $C = 5$ communities with respective intra- and inter-community edge probabilities of $0.8$ and $0.2$. The initial graph signal $\bbdelta_i \in \reals^N$ is a Kronecker delta centered at node $i$ and its realization at time $t$ is $\bbx_t = \bbS^{t} \bbdelta_{i}$ with $\bbS = \bbW / \lambda_{\max}(\bbW)$. We generated the training set $\{\bbx,i\}$ comprising $10,000$ samples by selecting uniformly at random both $i \!\in\! \{1,\ldots,C\}$ and $t \!\in\! \{0,\ldots,N\}$. We then tested the different approaches on $200$ new samples and averaged the performance over $10$ different data and $10$ different graph realizations for an overall of $100$ Monte-Carlo runs.


\textbf{Models and results.} We considered seven architectures each of them composed of the cascade of a graph filtering layer with ReLU nonlinearity and a fully connected layer with softmax nonlinearity. The architectures are: $a)$ a spectral GCNN \eqref{eq:specGCNN} with $\bbB$ being a cubic spline kernel and $b = 5$; $b)$ a polynomial GCNN \eqref{eq:polFilt} of order $K = 4$; $c)$ two NV GNNs \eqref{eq:nodeV} of order $K = 4$ and $|\ccalB| = 5$ \emph{privileged} nodes selected by $b_{i})$ maximum degree and $b_{ii})$ spectral proxies \cite{anis2016efficient}; $d)$ an EV GNN \eqref{eq:evConv} of order $K = 4$; and $e)$ two HEV GNNs \eqref{eq:HEV} of order $K = 4$ and $|\ccalB| = 5$ \emph{privileged} nodes selected similarly to the NV case. We used the ADAM optimizer with a learning rate $\mu=0.001$ and decaying factors $\beta_{1}=0.9$ and $\beta_{2}=0.999$ run over $20$ epochs with batches of size $100$.

Table~\ref{table_sourceloc} shows the obtained results where we see that because of their increased capacity the EV and the HEV outperform the other alternatives. We observe that the hybrid approaches exploit better the \emph{edge-varying} part when $\ccalB$ is composed of the nodes with the highest degree.


\subsection{Authorship Attribution} \label{subsec:author}

\textbf{Setup.} In this experiment, we aim to classify if a text excerpt belongs to Edgar Allan Poe or to any other contemporary author. For each text excerpt, we built the graph from the word adjacency network (WAN) between function words that act as nodes. These WANs serve as stylistic signatures for the author (see \cite{segarra15-wans} for full details). Fixed then WAN for Poe, we treat the frequency count of the function words as a graph signal.


In particular, we considered $846$ text excerpts by Poe and randomly split the dataset into $608$ training, $68$ validation, and $170$ testing texts. We sumed the adjacency matrices of the WANs obtained from the $608$ training texts to get the Poe's signature graph. We completed the training, validation, and test sets by adding respectively other $608$, $68$, and $170$ randomly selected texts by contemporary authors. We set $\bbS$ to be the adjacency matrix of the Poe's signature graph and averaged the performance over $10$ different data splits.

\textbf{Models and results.} We analyzed the same architectures as in the previous section but set the number of output features to $F = 2$, the recursion orders to $K = 1$, $b = 2$ for the spectral GCNN, and $|\ccalB| = 2$. We used the same ADAM optimizer for training now over $80$ epochs with batch sizes of $100$ samples.


The results in Table~\ref{table_author} show that the hybrid approaches offer the best performance highlighting the potential of solutions that consider both edge-dependent and global coefficients. In fact, the polynomial model with global coefficients suffers the most in this experiment.


\begin{table}[t]
    \centering
        \caption{Average performance (and std. dev.) for the source localization problem.}
    \begin{tabular}{lr} \hline
        Model 				& Accuracy 				\\ \hline
        Spectral    		        & \texttt{26.89($\pm$ 0.87)\%}	\\
        Polynomial      			& \texttt{74.55($\pm$ 7.32)\%}	\\
        Node Variant (NV) Degree    & \texttt{74.77($\pm$ 7.77)\%}	\\
        Node Variant (NV) S. Proxies& \texttt{75.62($\pm$ 8.19)\%}	\\
        Edge Variant (EV)			& \texttt{85.47($\pm$10.77)\%}	\\
        Hybrid EV (HEV) Degree		& \texttt{80.53($\pm$10.21)\%}	\\
        Hybrid EV (HEV) S. Proxies	& \texttt{75.37($\pm$ 8.20)\%}	\\ \hline
    \end{tabular}\vskip-.5cm
    \label{table_sourceloc}
\end{table}

\section{Conclusion}
\label{sec:concl}

We proposed a general framework that unifies $11$ state-of-the-art GCNN architectures into one recursion, named the edge-variant recursion. This unification highlighted the different tradeoff between the number of parameters and the amount of local detail that each approach adopts. Moreover, it shows rigorous ways to choose different tradeoffs and come up with a novel and ad hoc architecture for a problem at hand that is implemented locally in the vertex domain. We here proposed one, among many, extension and showed that it outperforms current solutions for graph signal classification tasks.

\begin{table}[t]
    \centering
        \caption{Average performance (and std. dev.) for the authorship attribution problem on text excerpts by Edgar Allan Poe.}
    \begin{tabular}{lr} \hline
        Model 				& Accuracy 				\\ \hline
        Spectral    		        & \texttt{88.88($\pm$ 1.50)\%}	\\
        Polynomial      			& \texttt{79.88($\pm$15.31)\%}	\\
        Node Variant (NV) Degree    & \texttt{88.88($\pm$ 2.62)\%}	\\
        Node Variant (NV) S. Proxies& \texttt{86.12($\pm$ 5.94)\%}	\\
        Edge Variant (EV)			& \texttt{89.00($\pm$ 2.11)\%}	\\
        Hybrid EV (HEV) Degree		& \texttt{89.18($\pm$ 1.99)\%}	\\
        Hybrid EV (HEV) S. Proxies	& \texttt{90.00($\pm$ 1.21)\%}	\\ \hline
    \end{tabular}\vskip-.5cm
    \label{table_author}
\end{table}

\bibliographystyle{IEEEtran}
\bibliography{myIEEEabrv,dissertation}

\end{document}